\documentclass{article}

\PassOptionsToPackage{numbers, compress}{natbib}

\usepackage[final]{neurips_2020}

\usepackage[utf8]{inputenc} 
\usepackage[T1]{fontenc}    
\usepackage{hyperref}       
\usepackage{url}            
\usepackage{booktabs}       
\usepackage{amsfonts}       
\usepackage{nicefrac}       
\usepackage{microtype}      

\hypersetup{hidelinks}

\usepackage{booktabs}       
\usepackage{amsfonts}       
\usepackage{microtype}      
\usepackage{url}
\usepackage{verbatim} 
\usepackage{graphicx}
\usepackage{multirow}
\usepackage{algorithm}
\usepackage[noend]{algorithmic}
\usepackage{xspace}
\usepackage{epsfig}
\usepackage{amsmath}
\usepackage{amsthm}
\usepackage{amssymb}
\usepackage{times}
\usepackage{xr}
\usepackage{bbm}
\usepackage{bm}
\usepackage{subcaption}
\usepackage{enumitem}
\usepackage{hyperref}       
\usepackage{multicol}
\usepackage{tabularx}
\usepackage{perpage} 
\MakePerPage{footnote} 
\usepackage{wrapfig}
\usepackage{floatflt}

\newcommand{\xhdr}[1]{{\noindent\bfseries #1}.}

\newcommand{\cut}[1]{}

\newcommand{\ie}{\emph{i.e.}}

\newcommand{\name}{\textsc{Grape}\xspace}

\usepackage{amsmath,amsfonts,bm}

\def\eqref#1{equation~\ref{#1}}

\def\1{\bm{1}}

\DeclareMathAlphabet{\mathsfit}{\encodingdefault}{\sfdefault}{m}{sl}
\SetMathAlphabet{\mathsfit}{bold}{\encodingdefault}{\sfdefault}{bx}{n}

\newcommand*{\affmark}[1][*]{\textsuperscript{#1}}
\newcommand*\samethanks[1][\value{footnote}]{\footnotemark[#1]}

\usepackage{cleveref}

\title{Handling Missing Data with \\Graph Representation Learning}

\author{
Jiaxuan You\affmark[1]\thanks{Equal contribution}\\
\And 
Xiaobai Ma\affmark[2]\samethanks\\
\And 
Daisy Yi Ding\affmark[3]\samethanks\\
\And 
Mykel Kochenderfer\affmark[2]
\And
Jure Leskovec\affmark[1]\\
}

\begin{document}

\maketitle
\vspace{-3em}
 \begin{center}
 \affmark[1]Department of Computer Science, \affmark[2]Department of Aeronautics and Astronautics, \\ and \affmark[3]Department of Biomedical Data Science, Stanford University \\ 
 \texttt{\{jiaxuan, jure\}@cs.stanford.edu\\ \{maxiaoba, dingd, mykel\}@stanford.edu}
 \end{center}

\begin{abstract}
Machine learning with missing data has been approached in two different ways, including \emph{feature imputation} where missing feature values are estimated based on observed values and \emph{label prediction} where downstream labels are learned directly from incomplete data.
However, existing imputation models tend to have strong prior assumptions and cannot learn from downstream tasks,
while models targeting label prediction often involve heuristics and can encounter scalability issues.
Here we propose \name, a graph-based framework for feature imputation as well as label prediction. \name tackles the missing data problem using a \emph{graph representation}, where the observations and features are viewed as two types of nodes in a bipartite graph, and the observed feature values as edges.
Under the \name framework,
the \emph{feature imputation} is formulated as an \emph{edge-level prediction} task and the \emph{label prediction} as a \emph{node-level prediction} task. 
These tasks are then solved with Graph Neural Networks.
Experimental results on nine benchmark datasets show that \name yields 20\% lower mean absolute error for imputation tasks and 10\% lower for label prediction tasks, compared with existing state-of-the-art methods.
\end{abstract}

\section{Introduction}

\begin{figure}[t]
\centering
\includegraphics[width=\columnwidth]{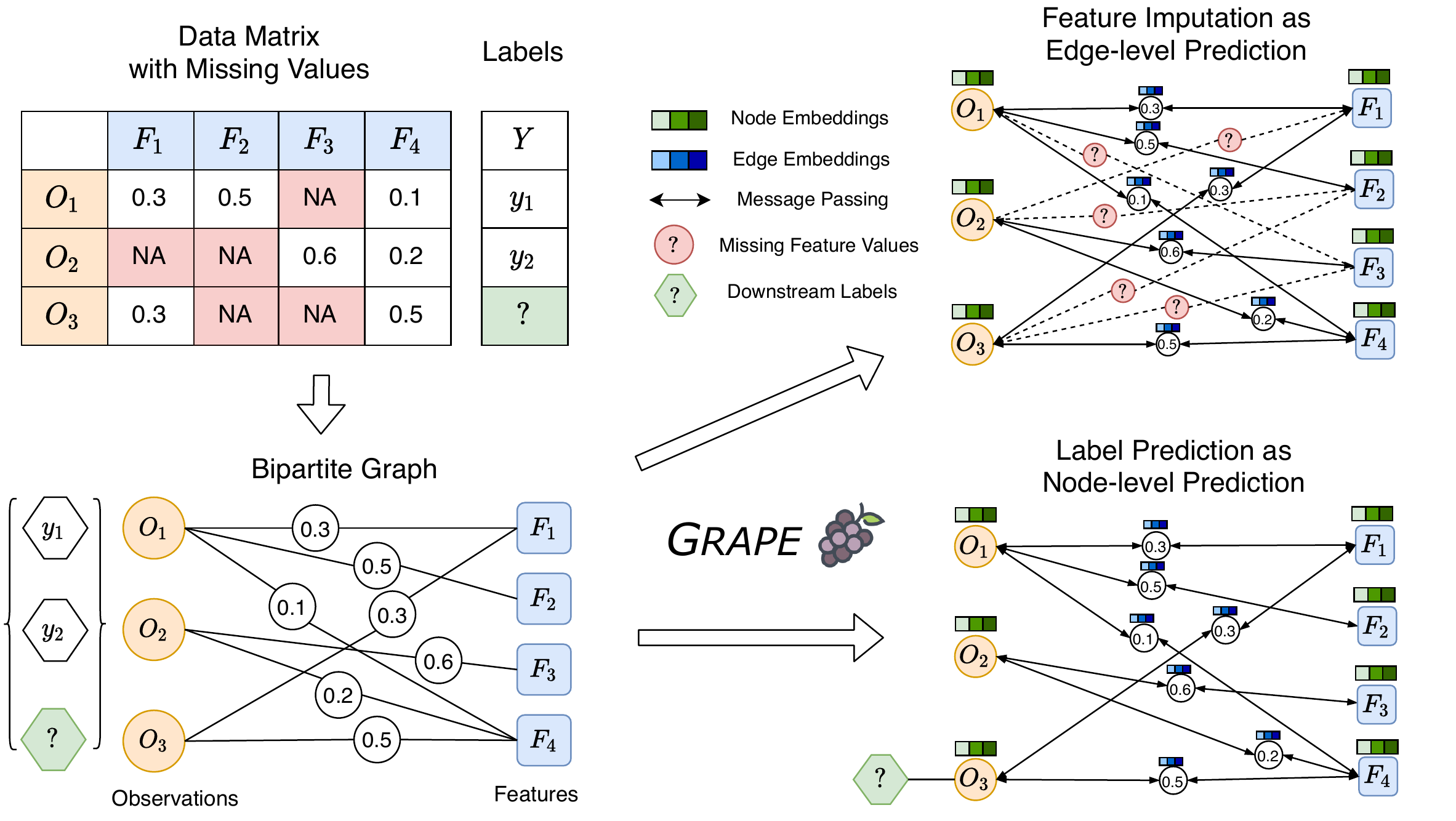}
\caption{
In the \name framework, we construct a bipartite graph from the data matrix with missing feature values, where the entries of the matrix in red indicate the missing values (\textbf{Top Left}). To construct the graph, the observations $O$ and features $F$ are considered as two types of nodes and the observed values in the data matrix are viewed as weighted/attributed edges between the observation and feature nodes (\textbf{Bottom Left}). With the constructed graph, we formulate the feature imputation problem and the label prediction problem as edge-level (\textbf{Top right}) and node-level (\textbf{Bottom right}) prediction tasks, respectively. The tasks can then be solved with our \name GNN model that learns node and edge embeddings through rounds of message passing.
}
\label{fig:graph}
\end{figure}

Issues with learning from incomplete data arise in many domains including computational biology, clinical studies, survey research, finance, and economics \cite{brick1996handling, little2019statistical, sterne2009multiple, troyanskaya2001missing, wooldridge2007inverse}.
The missing data problem has previously been approached in two different ways: \emph{feature imputation} and \emph{label prediction}.
Feature imputation 
involves estimating missing feature values based on observed values \cite{cai2010singular, candes2009exact, dempster1977maximum, garcia2010pattern,  ghahramani1994supervised, gondara2017multiple, hastie2015matrix, mazumder2010spectral,  srebro2005maximum,  stekhoven2012missforest, troyanskaya2001missing,   van2007multiple, buuren2010mice, vincent2008extracting,  yoon2018gain}, and 
label prediction aims to directly accomplish a downstream task, such as classification or regression, with the missing values present in the input data \cite{bengio1996recurrent, breiman1984classification,  chechik2008max,  ghahramani1994supervised, goldberg2010transduction, hazan2015classification, pelckmans2005handling, shivaswamy2006second,  smieja2018processing, williams2005incomplete, xia2017adjusted}.

Statistical methods for feature imputation often provide useful theoretical properties but exhibit notable shortcomings:
(1) they tend to make strong assumptions about the data distribution;
(2) they lack the flexibility for handling mixed data types that include both continuous and categorical variables;
(3) matrix completion based approaches cannot generalize to unseen samples and require retraining when the model encounters new data samples \cite{cai2010singular, candes2009exact, hastie2015matrix, mazumder2010spectral,  srebro2005maximum,  troyanskaya2001missing}.
When it comes to models for label prediction, existing approaches such as tree-based methods rely on heuristics \cite{breiman1984classification} and tend to have scalability issues.
For instance, one of the most popular procedures called surrogate splitting does not scale well, because each time an original splitting variable is missing for some observation it needs to rank all other variables as surrogate candidates and select the best alternative. 

Recent advances in deep learning have enabled new approaches to handle missing data.
Existing imputation approaches often use deep generative models, such as Generative Adversarial Networks (GANs) \cite{yoon2018gain} or autoencoders \cite{gondara2017multiple, vincent2008extracting}, to reconstruct missing values.
While these models are flexible, they have several limitations:
(1) when imputing missing feature values for a given observation, these models fail to make full use of feature values from other observations;
(2) they tend to make biased assumptions about the missing values by initializing them with special default values.

Here, we propose \name \footnote{Project website with data and code: \url{http://snap.stanford.edu/grape}}, a general framework for feature imputation and label prediction in the presence of missing data. 
Our key innovation is to formulate the problem using a \emph{graph representation}, where we construct a bipartite graph with observations and features as two types of nodes, and the observed feature values as attributed edges between the observation and feature nodes (\Cref{fig:graph}). 
Under this graph representation, the \emph{feature imputation} can then be naturally formulated as an \emph{edge-level prediction} task, and the \emph{label prediction} as a \emph{node-level prediction} task. 

\name solves both tasks via Graph Neural Networks (GNNs). 
Specifically, \name adopts a GNN architecture inspired by the GraphSAGE model \cite{hamilton2017inductive}, while having three innovations in its design: 
(1) since the edges in the graph are constructed based on the data matrix and have rich attribute information, we introduce \emph{edge embeddings} during message passing and incorporate both discrete and continuous edge features in the message computation; 
(2) we design \emph{augmented node features} to initialize observation and feature nodes, which provides greater representation power and maintains inductive learning capabilities;
(3) to overcome the common issue of overfitting in the missing data problem, we employ an \emph{edge dropout} technique that greatly boosts the performance of \name.

We compare \name with the state-of-the-art feature imputation and label prediction algorithms on 9 benchmark datasets from the UCI Machine Learning Repository \cite{uci}.
In particular, \name yields 20\% lower mean absolute error (MAE) for the imputation tasks and 10\% lower MAE for the prediction tasks at the 30\% data missing rate. 
Finally, we demonstrate \name's strong generalization ability by showing its superior performance on unseen observations without the need for retraining.

Overall, our approach has several important benefits: (1) by creating a bipartite graph structure we create connections between different features (via observations) and similarly between the observations (via features);
(2) GNN elegantly harnesses this structure by learning to propagate and borrow information from other features/observations in a graph localized way; (3) GNN allows us to model both feature imputation as well as label prediction in an end-to-end fashion, which as we show in experiments leads to strong performance improvements.

\section{Related Work}

\xhdr{Feature imputation}
Successful statistical approaches for imputation include joint modeling with  Expectation-Maximization \cite{dempster1977maximum,garcia2010pattern, ghahramani1994supervised, honaker2011amelia},
multivariate imputation by chained equations (MICE) \cite{burgette2010multiple, raghunathan2001multivariate, stekhoven2012missforest, van2007multiple, buuren2010mice}, $k$-nearest neighbors (KNN) \cite{kim2004reuse,troyanskaya2001missing}, and matrix completion \cite{cai2010singular, candes2009exact, hastie2015matrix,  mazumder2010spectral, srebro2005maximum,  troyanskaya2001missing}.
However, joint modeling tends to make assumptions about the data distribution through a parametric density function; joint modeling and matrix completion lack the flexibility to handle data of mixed modalities; MICE and KNN cannot accomplish imputation while adapting to downstream tasks. 

Recently, deep learning models have also been used to tackle
the feature imputation problem \cite{gondara2017multiple, spinelli2019missing, vincent2008extracting,  yoon2018gain}.
However, these models have important limitations.
Denoising autoencoder (DAE) models \cite{gondara2017multiple, vincent2008extracting} and
GAIN \cite{yoon2018gain} only use a single observation as input to impute the missing features. In contrast, \name explicitly captures the complex interactions between multiple observations and features.
GNN-based approaches have also been proposed in the context of matrix completion \cite{gnn_matrix_completion, hartford2018deep, monti2017geometric, zhang2019inductive, zheng2018spectral}.
However, they often make the assumption of finite, known-range values in their model design, which limits their applicability to imputation problems with continuous values.
In contrast, \name can handle both continuous and discrete feature values. 

\xhdr{Label prediction with the presence of missing data}
Various models have been adapted for label prediction with the presence of missing data, including tree-based approaches  \cite{breiman1984classification, xia2017adjusted}, probabilistic modeling \cite{ghahramani1994supervised}, logistic regression \cite{williams2005incomplete}, support vector machines \cite{chechik2008max, pelckmans2005handling}, deep learning-based models \cite{bengio1996recurrent, goodfellow2013multi, smieja2018processing}, and many others \cite{goldberg2010transduction, hazan2015classification, liao2007quadratically, shivaswamy2006second}.
Specifically, decision tree is a classical statistical approach that can handle missing values for the label prediction task \cite{breiman1984classification}.
With the surrogate splitting procedure, decision tree uses a single surrogate variable to replace the original splitting variable with missing values, which is effective but inefficient,
and has been shown to be inferior to the ``impute and then predict'' procedure \cite{feelders1999handling}. 
Random forests further suffer from the scalability issues as they consist of multiple decision trees \cite{liaw2002classification, xia2017adjusted}.
In contrast, \name handles the missing feature entries naturally with the \emph{graph representation} without any additional heuristics. The computation of \name is efficient and easily parallelizable with modern deep learning frameworks. 

\xhdr{Overall discussion}
In \name implementation, we adopt several successful GNN design principles. Concretely, our core architecture is inspired by GraphSAGE \cite{hamilton2017inductive}; we apply GraphSAGE to bipartite graphs following G2SAT \cite{you2019g2sat}; we use edge dropout in~\cite{rong2019dropedge}; we use one-hot auxiliary node features which has been used in \cite{murphy2019relational,you2019position}; we follow the GNN design guidelines in \cite{you2020design} to select hyperparameters. Moreover, matrix completion tasks have been formulated as bipartite graphs and solved via GNNs in \cite{gnn_matrix_completion,zhang2019inductive}; however, they only consider the feature imputation task with discrete feature values.
We emphasize that our main contribution is \emph{not the particular GNN model but the graph-based framework for the general missing data problem}. \name is the first graph-based solution to both feature imputation and label prediction aspects of the missing data problem.

\section{The \name Framework}

\subsection{Problem Definition}
\label{subsec:problemdefinition}
Let $\mathbf{D}\in \mathbb{R}^{n\times m}$ be a feature matrix consisting of $n$ data points and $m$ features. The $j$-th feature of the $i$-th data point is denoted as $\mathbf{D}_{ij}$. In the missing data problem, certain feature values are missing, denoted as a mask matrix $\mathbf{M}\in \{0,1\}^{n\times m}$ where the value of $\mathbf{D}_{ij}$ can be observed only if $\mathbf{M}_{ij}=1$. 
Usually, datasets come with labels of a downstream task.
Let $\mathbf{Y}\in \mathbb{R}^n$ be the label for a downstream task and $\mathbf{V}\in \{0,1\}^n$ the train/test partition, where $\mathbf{Y}_i$  can be observed at training test only if $\mathbf{V}_i=1$.
We consider two tasks: (1) feature imputation, where the goal is to predict the missing feature values $\mathbf{D}_{ij}$ at $\mathbf{M}_{ij}=0$; (2) label prediction, where the goal is to predict test labels $\mathbf{Y}_i$ at $\mathbf{V}_i=0$.

\subsection{Missing Data Problem as a Graph Prediction Task}
\label{subsec:mdi_to_graph}
The key insight of this paper is to represent the feature matrix with missing values as a \emph{bipartite graph}. Then the feature imputation problem and the label prediction problem can naturally be formulated as node prediction and edge prediction tasks (Figure~\ref{fig:graph}).

\xhdr{Feature matrix as a bipartite graph}
The feature matrix $\mathbf{D}$ and the mask $\mathbf{M}$ can be represented as an undirected bipartite graph $\mathcal{G}=(\mathcal{V}, \mathcal{E})$, where $\mathcal{V}$ is the node set that consists of two types of nodes $\mathcal{V}=\mathcal{V}_D \cup \mathcal{V}_F$, $\mathcal{V}_D=\{u_1,...,u_n\}$ and $\mathcal{V}_F=\{v_1, \ldots, v_m\}$, $\mathcal{E}$ is the edge set where edges only exist between nodes in different partitions: $\mathcal{E}=\{(u_i,v_j,\mathbf{e}_{u_iv_j}) \mid u_i\in \mathcal{V}_D, v_j\in \mathcal{V}_F, \mathbf{M}_{ij}=1\}$, where the edge feature, $\mathbf{e}_{u_iv_j}$, takes the value of the corresponding feature $\mathbf{e}_{u_iv_j}=\mathbf{D}_{ij}$. If $\mathbf{D}_{ij}$ is a discrete variable then it is transformed to a one-hot vector then assigned to $\mathbf{e}_{u_iv_j}$. To simplify the notation $\mathbf{e}_{u_iv_j}$, we use $\mathbf{e}_{ij}$ in the context of feature matrix $\mathbf{D}$, and $\mathbf{e}_{uv}$ in the context of graph $\mathcal{G}$.

\xhdr{Feature imputation as edge-level prediction}
Using the definitions above, imputing missing features can be represented as learning the edge  value prediction mapping: $\hat{\mathbf{D}}_{ij} = \hat{\mathbf{e}}_{ij} = f_{ij}(\mathcal{G})$ by minimizing the difference between $\hat{\mathbf{D}}_{ij}$ and $\mathbf{D}_{ij},\forall \mathbf{M}_{ij}=0$.
When imputing discrete attributes, we use cross entropy loss. When imputing continuous values, we use MSE loss.

\xhdr{Label prediction as node-level prediction}
Predicting downstream node labels can be represented as learning the mapping: $\hat{\mathbf{Y}}_i=g_i(\mathcal{G})$ by minimizing the difference between $\hat{\mathbf{Y}}_i$ and $\mathbf{Y}_i,\forall \mathbf{V}_i=0$. 

\subsection{Learning with \name}
\label{subsec:network}

\name adopts a GNN architecture inspired by GraphSAGE~\cite{hamilton2017inductive}, which is a variant of GNNs that has been shown to have strong inductive learning capabilities across different graphs. 
We extend GraphSAGE to a bipartite graph setting by adding multiple important components that ensure its successful application to the missing data problem.

\xhdr{\name GNN architecture}
Given that our bipartite graph $\mathcal{G}$ has important information on its edges, we modify GraphSAGE architecture by introducing \emph{edge embeddings}. At each GNN layer $l$, the message passing function takes the concatenation of the embedding of the source node $\mathbf{h}_v^{(l-1)}$ and the edge embedding $\mathbf{e}_{uv}^{(l-1)}$ as the input:
\begin{equation}
    \mathbf{n}_v^{(l)} = \textsc{Agg}_l\Big(\sigma(\mathbf{P}^{(l)} \cdot \textsc{Concat}( \mathbf{h}_v^{(l-1)}, \mathbf{e}_{uv}^{(l-1)})\mid \forall u\in \mathcal{N}(v,\mathcal{E}_{drop}))\Big)
\end{equation}
where $\textsc{Agg}_l$ is the aggregation function, $\sigma$ is the non-linearity, $\mathbf{P}^{(l)}$ is the trainable weight, $\mathcal{N}$ is the node neighborhood function.
Node embedding $\mathbf{h}_v^{(l)}$ is then updated using:
\begin{equation}
\mathbf{h}_v^{(l)} = \sigma(\mathbf{Q}^{(l)} \cdot \textsc{Concat}(\mathbf{h}_v^{(l-1)}, \mathbf{n}_v^{(l)}))
\end{equation}
where $\mathbf{Q}^{(l)}$ is the trainable weight,
we additionally update the edge embedding $\mathbf{e}_{uv}^{(l)}$ by:
\begin{equation}
    \mathbf{e}_{uv}^{(l)} = \sigma(\mathbf{W}^{(l)} \cdot \textsc{Concat}(\mathbf{e}_{uv}^{(l-1)}, \mathbf{h}_u^{(l)}, \mathbf{h}_v^{(l)}))
\end{equation}
where $\mathbf{W}^{(l)}$ is the trainable weight.
To make edge level predictions at the $L$-th layer:
\begin{equation}
    \label{eq:edge_pred}
    \hat{\mathbf{D}}_{uv} = \mathbf{O}_{edge}( \textsc{Concat}(\mathbf{h}_u^{(L)}, \mathbf{h}_v^{(L)}))
\end{equation}
The node-level prediction is made using the imputed dataset $\hat{\mathbf{D}}$:
\begin{equation}
    \label{eq:node_pred}
    \hat{\mathbf{Y}}_u = \mathbf{O}_{node}(\hat{\mathbf{D}}_{u\cdot})
\end{equation}
where $\mathbf{O}_{edge}$ and $\mathbf{O}_{node}$ are feedforward neural networks.

\xhdr{Augmented node features for bipartite message passing}
Based on our definition, nodes in $\mathcal{V}_D$ and $\mathcal{V}_F$ do not naturally come with features. The straightforward approach would be to augment nodes with constant features.
However, such formulation would make \name hard to differentiate messages from different feature nodes in $\mathcal{V}_F$.
In real-world applications, different features can represent drastically different semantics or modalities. 
For example in the \emph{Boston Housing} dataset from UCI \cite{uci}, some features are categorical such as if the house is by the Charles River, while others are continuous such as the size of the house. 

Instead, we propose to use $m$-dimensional one-hot node features for each node in $\mathcal{V}_F$ ($m = |\mathcal{V}_F|$), while using $m$-dimensional\footnote{We make data nodes and feature nodes to have the same feature dimension for the ease of implementation.} constant vectors as node feature for data nodes in $\mathcal{V}_F$:
\begin{equation}
    \textsc{Init}(v) = \begin{cases}
    \mathbf{1} & v \in \mathcal{V}_D\\
    \textsc{OneHot} &  v \in \mathcal{V}_F
\end{cases}
\end{equation}

Such a formulation leads to a better representational power to differentiate feature nodes with different underlying semantics or modalities.
Additionally, the formulation has the capability of generalizing the trained \name to completely unseen data points in the given dataset.
Furthermore, it allows us to transfer knowledge from an external dataset with the same set of features to the dataset of interest, which is particularly useful when the external dataset provides rich information on the interaction between observations and features (as captured by \name).
For example, as a real-world application in biomedicine, gene expression data can be used to predict disease types and frequently contain missing values. 
If we aim to impute missing values in a gene expression dataset of a small cohort of lung cancer patients, public datasets, e.g., the Cancer Genome Atlas Program (TCGA) \cite{weinstein2013cancer} can be first leveraged to train \name, where rich interactions between patients and features are learned.
Then, the trained \name can be applied to our smaller dataset of interest to accomplish imputation.

\xhdr{Improved model generalization with edge dropout}
When doing feature imputation, a naive way of training \name is to directly feed $\mathcal{G}=(\mathcal{V}; \mathcal{E})$ as the input. However, since all the observed edge values are used as the input, an identity mapping $\hat{\mathbf{D}}_{ij}=\mathbf{e}^{(0)}_{ij}$ is enough to minimize the training loss; therefore, \name trained under this setting easily overfits the training set. To force the model to generalize to unseen edge values, we randomly mask out edges $\mathcal{E}$ with dropout rate $r_{drop}$:
\begin{equation}
    \textsc{DropEdge}(\mathcal{E},r_{drop})=\{(u_i,v_j,\mathbb{e}_{ij}) \mid (u_i,v_j,\mathbf{e}_{ij})\in \mathcal{E},\mathbf{M}_{drop,ij}>r_{drop}\}
\end{equation}
where $\mathbf{M}_{drop}\in \mathbb{R}^{n\times m}$ is a random matrix sampled uniformly in $(0,1)$.
This approach is similar to DropEdge~\cite{rong2019dropedge}, but with a more direct motivation for feature imputation. At test time, we feed the full graph $\mathcal{G}$ to \name.
Overall, the complete computation of \name is summarized in~\Cref{alg:train}.

\setlength{\textfloatsep}{3mm}
\begin{algorithm}[t]
\caption{\name forward computation}
\label{alg:train}
\textbf{Input:} Graph $\mathcal{G}=(\mathcal{V}; \mathcal{E})$; Number of layers $L$; Edge dropout rate $r_{drop}$;
Weight matrices $\mathbf{P}^{(l)}$ for \textit{message passing}, $\mathbf{Q}^{(l)}$ for \textit{node updating}, and $\mathbf{W}^{(l)}$ for \textit{edge updating}; non-linearity $\sigma$; aggregation functions $\textsc{Agg}_l$; neighborhood function $\mathcal{N}:v \times \mathcal{E} \rightarrow 2^\mathcal{V}$\\
\textbf{Output:} Node embeddings $\mathbf{h}_v$ corresponding to each $v \in \mathcal{V}$ 
\begin{algorithmic}[1]
\STATE $\mathbf{h}_{v}^{(0)} \leftarrow \textsc{Init}(v),\forall v \in \mathcal{V}$
\STATE $\mathbf{e}_{uv}^{(0)} \leftarrow \mathbf{e}_{uv},\forall \mathbf{e}_{uv} \in \mathcal{E}$
\STATE $\mathcal{E}_{drop} \leftarrow \textsc{DropEdge}(\mathcal{E},r_{drop})$
\FOR{$l\in \{1,\dots,L\}$}

\FOR{$v \in \mathcal{V}$}
    \STATE $\mathbf{n}_v^{(l)} = \textsc{Agg}_l\Big(\sigma(\mathbf{P}^{(l)} \cdot \textsc{Concat}( \mathbf{h}_v^{(l-1)}, \mathbf{e}_{uv}^{(l-1)})\mid \forall u\in \mathcal{N}(v,\mathcal{E}_{drop}))\Big)$
    
    \STATE $\mathbf{h}_v^{(l)} = \sigma(\mathbf{Q}^{(l)} \cdot \textsc{Concat}(\mathbf{h}_v^{(l-1)}, \mathbf{n}_v^{(l)}))$
    
\ENDFOR

\FOR{$(u,v) \in \mathcal{E}_{drop}$}
    \STATE $\mathbf{e}_{uv}^{(l)} = \sigma(\mathbf{W}^{(l)} \cdot \textsc{Concat}(\mathbf{e}_{uv}^{(l-1)}, \mathbf{h}_u^{(l)}, \mathbf{h}_v^{(l)}))$
\ENDFOR

\STATE $z_v \leftarrow h_v^L$  
\ENDFOR
\end{algorithmic}

\end{algorithm}

\section{Experiments}
\label{sec:exp}

\subsection{Experimental Setup}

\xhdr{Datasets}
We conduct experiments on 9 datasets from the UCI Machine Learning Repository \cite{uci}. 
The datasets come from different domains including civil engineering (\textsc{concrete}, \textsc{energy}), biology (\textsc{protein}), thermal dynamics (\textsc{naval}), etc.
The smallest dataset (\textsc{yacht}) has 314 observations and 6 features, while the largest dataset (\textsc{protein}) has over 45,000 observations and 9 features.
The datasets are fully observed; therefore, we introduce missing values by randomly removing values in the data matrix.
The attribute values are scaled to $[0,1]$ with a MinMax scaler \cite{leskovecmining}.

\xhdr{Baseline models}
\label{subsec:baseline}
We compare our model against five commonly used imputation methods. We also compare with a state-of-the-art deep learning based imputation model as well as a decision tree based label prediction model. More details on the baseline models are provided in the Appendix.
\begin{itemize}[noitemsep,topsep=0pt,leftmargin=5mm]
    \item Mean imputation (Mean): The method imputes the missing $\mathbf{D}_{ij}$ with the mean of all the samples with observed values in dimension $j$.
    \item K-nearest neighbors (KNN): The method imputes the missing value $\mathbf{D}_{ij}$ using the KNNs that have observed values in dimension $j$ with weights based on the Euclidean distance to sample $i$.
    \item Multivariate imputation by chained equations (MICE): The method runs multiple regression where each missing value is modeled conditioned on the observed non-missing values.
    \item Iterative SVD (SVD) \cite{troyanskaya2001missing}: The method imputes missing values based on matrix completion with iterative low-rank SVD decomposition.
    \item Spectral regularization algorithm (Spectral)~\cite{mazumder2010spectral}: This matrix completion model uses the nuclear norm as a regularizer and  imputes missing values with iterative soft-thresholded SVD.
    \item GAIN~\cite{yoon2018gain}, state-of-the-art deep imputation model with generative adversarial training~\cite{gan}.
    \item Decision tree (Tree) \cite{breiman1984classification}, a commonly used statistical method that can handle missing values for label prediction. We consider this baseline only for the label prediction task.\footnote{
    Random forest is not included due to the lack of a public implementation that can handle missing data without imputation.}
\end{itemize}

\begin{figure}[t]
\centering
\includegraphics[width=0.85\columnwidth]{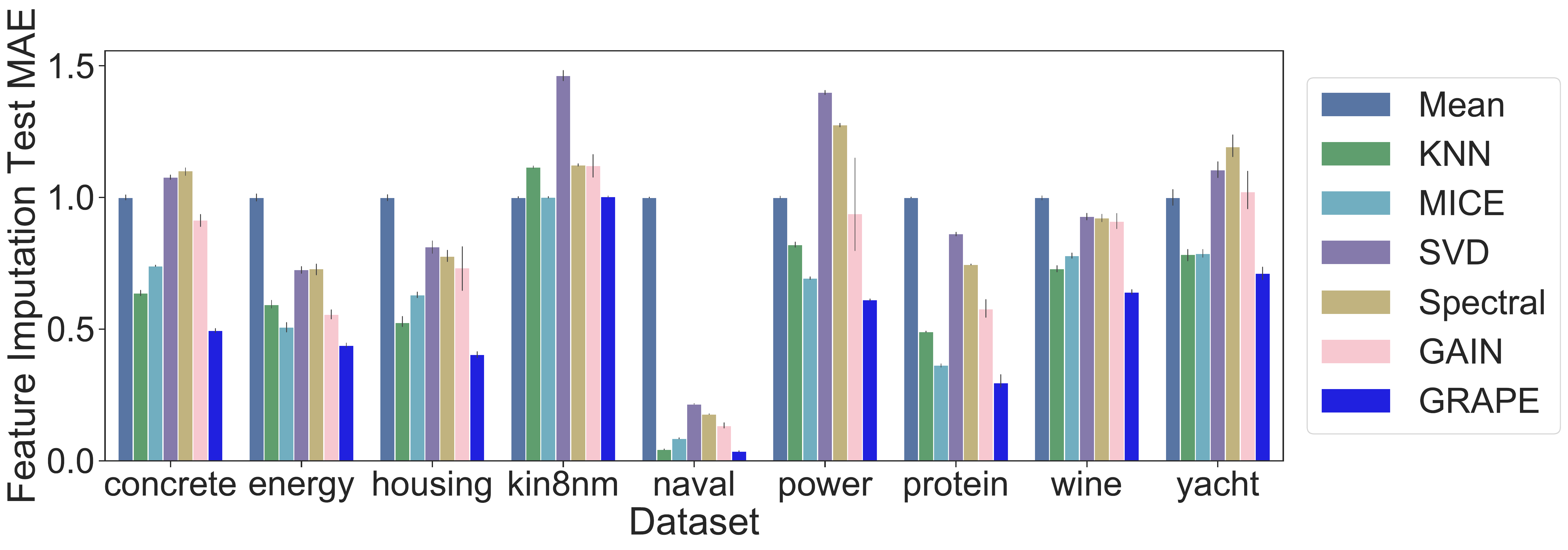}
\includegraphics[width=0.85\columnwidth]{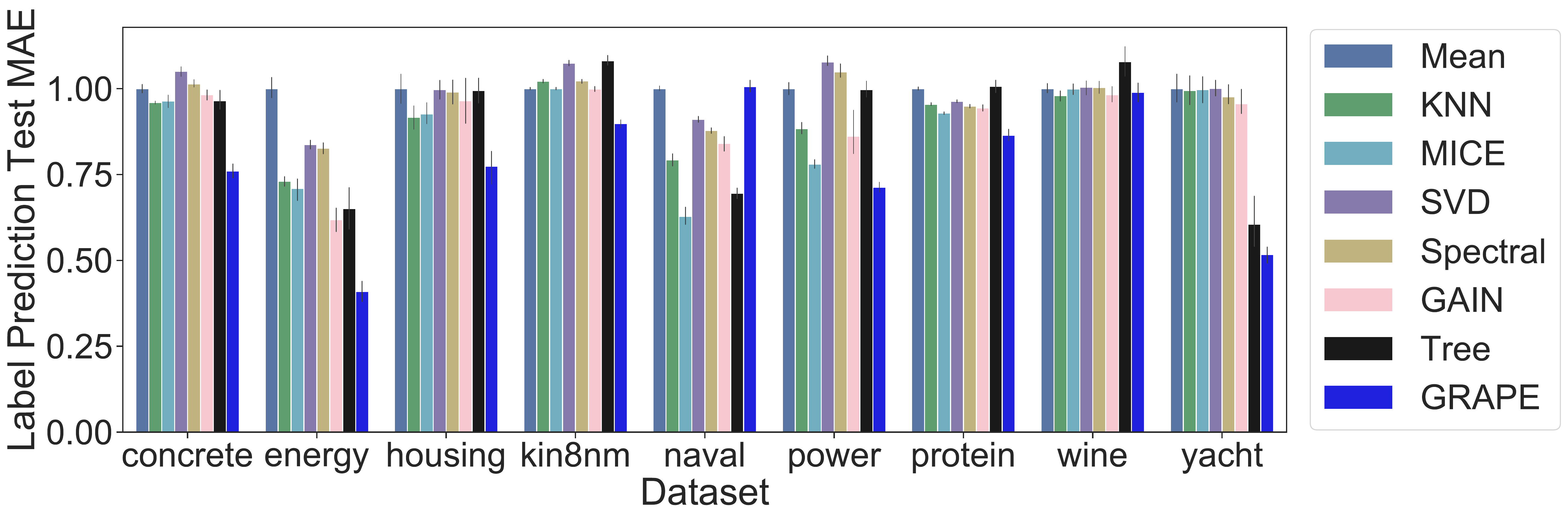}
\caption{Averaged MAE of \emph{feature imputation} (upper) and \emph{label prediction} (lower) on UCI datasets over 5 trials at data missing level of 0.3. The result is normalized by the average performance of Mean imputation. 
\name yields 20\% lower MAE for imputation and 10\% lower MAE for prediction compared with the best baselines (KNN for imputation and MICE for prediction).
}
\vspace{3mm}
\label{fig:mdi_reg}
\end{figure}

\xhdr{\name configurations}
For all experiments, we train \name for 20,000 epochs using the Adam optimizer \cite{kingma2014adam} with a learning rate at 0.001. 
For all \emph{feature imputation} tasks, we use a 3-layer GNN with 64 hidden units and $\textsc{ReLU}$ activation. 
The $\textsc{AGG}_l$ is implemented as a mean pooling function $\textsc{Mean}(\cdot)$ and $\mathbf{O}_{edge}$ as a multi-layer perceptron (MLP) with 64 hidden units. 
For \emph{label prediction} tasks, we use two GNN layers with 16 hidden units. $\mathbf{O}_{edge}$ and $\mathbf{O}_{node}$ are implemented as linear layers.
The edge dropout rate is set to $r_{drop}=0.3$.
For all experiments, we run 5 trials with different random seeds and report the mean and standard deviation of the results. 

\subsection{Feature Imputation}
\label{subsec:uci_mdi}
\xhdr{Setup}
We first compare the feature imputation performance of \name and all other imputation baselines. Given a full data matrix $\mathbf{D}\in \mathbb{R}^{n\times m}$, we generate a random mask matrix $\mathbf{M}\in \{0,1\}^{n\times m}$ with $P(\mathbf{M}_{ij}=0)=r_{miss}$ at a data missing level $r_{miss} = 0.3$. 
A bipartite graph $\mathcal{G}=(\mathcal{V},\mathcal{E})$ is then constructed based on $\mathbf{D}$ and $\mathbf{M}$ as described in \Cref{subsec:mdi_to_graph}. $\mathcal{G}$ is used as the input to \name at both the training and test time. 
The training loss is defined as the mean squared error (MSE) between $\mathbf{D}_{ij}$ and $\hat{\mathbf{D}}_{ij}$, $\forall \mathbf{M}_{ij}=1$. 
The test metric is defined as the mean absolute error (MAE)  between $\mathbf{D}_{ij}$ and $\hat{\mathbf{D}}_{ij}$, $\forall \mathbf{M}_{ij}=0$. 

\xhdr{Results}
As shown in \Cref{fig:mdi_reg}, \name has the lowest MAE on all datasets and its average error is 20\% lower compared with the best baseline (KNN). 
Since there are significant differences between the characteristics of different datasets, statistical methods often need to adjust its hyper-parameters accordingly, such as the cluster number in KNN, the rank in SVD, and the sparsity in Spectral. On the contrary, \name is able to adjust its trainable parameters adaptively through loss backpropagation and learn different observation-feature relations for different datasets. Compared with GAIN, which uses an MLP as the generative model, the GNN used in \name is able to explicitly model the information propagation process for predicting missing feature values.

\begin{figure}[t]
\centering
\includegraphics[width=0.28\columnwidth]{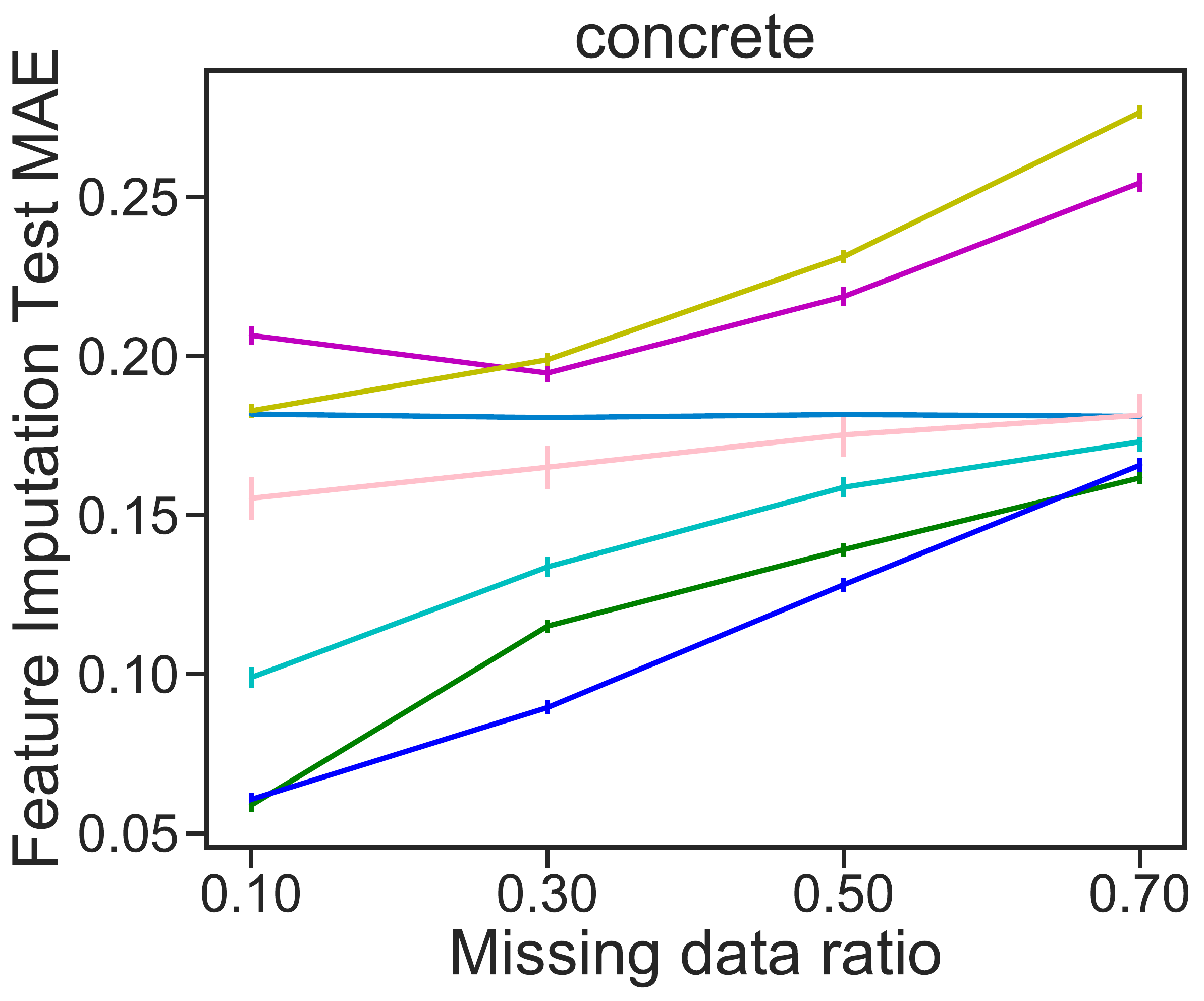}
\includegraphics[width=0.27\columnwidth]{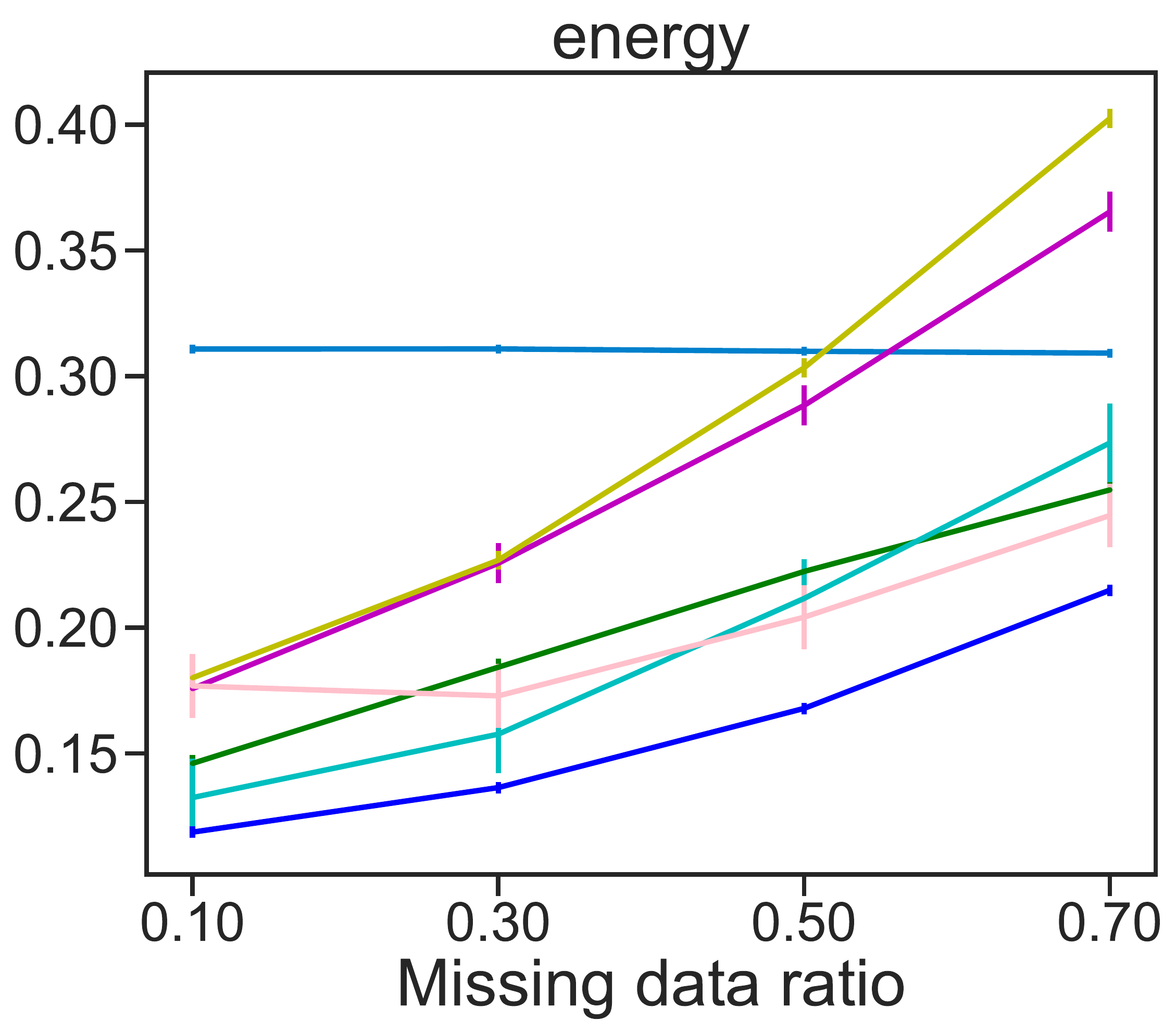}
\includegraphics[width=0.36\columnwidth]{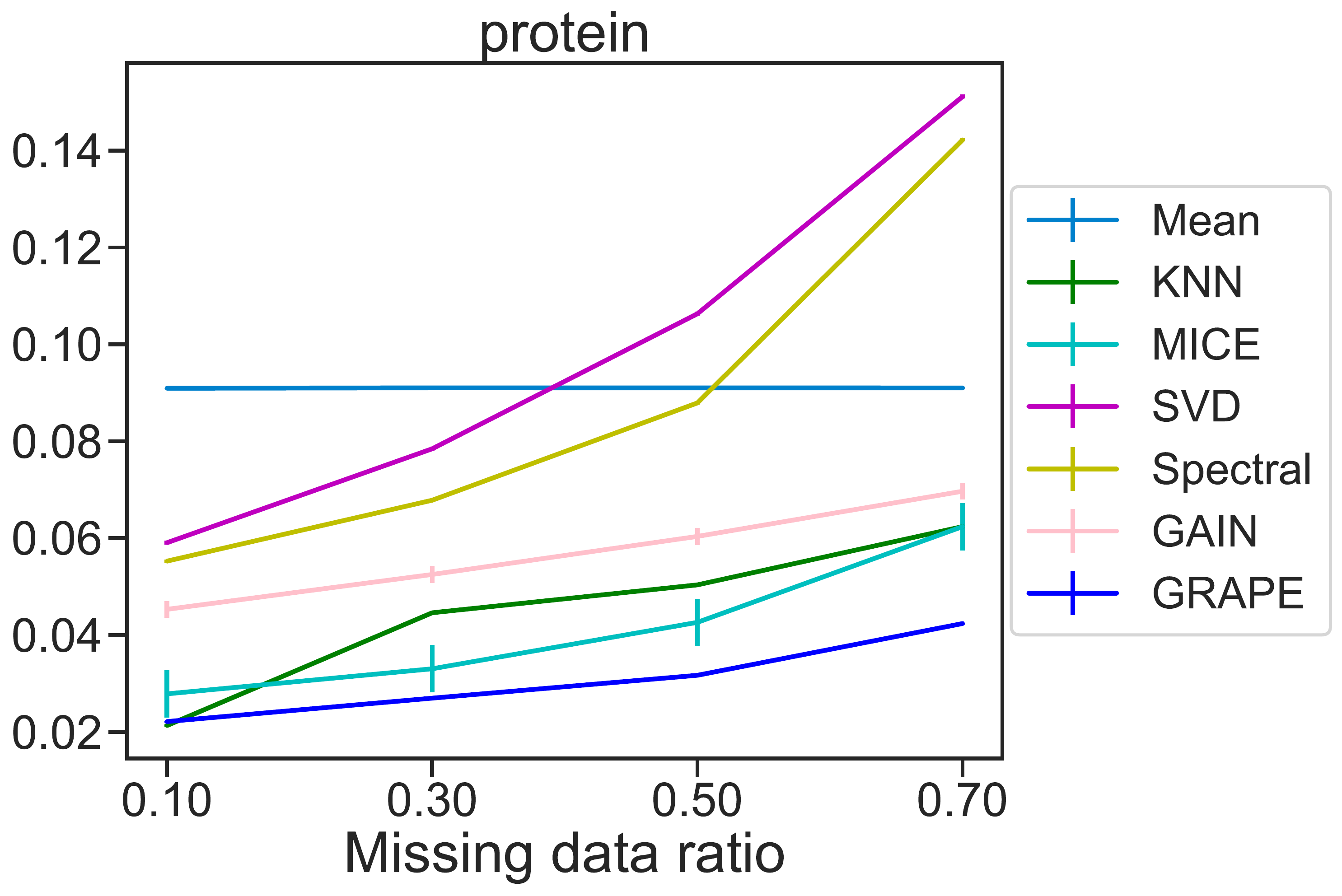}

\includegraphics[width=0.3\columnwidth]{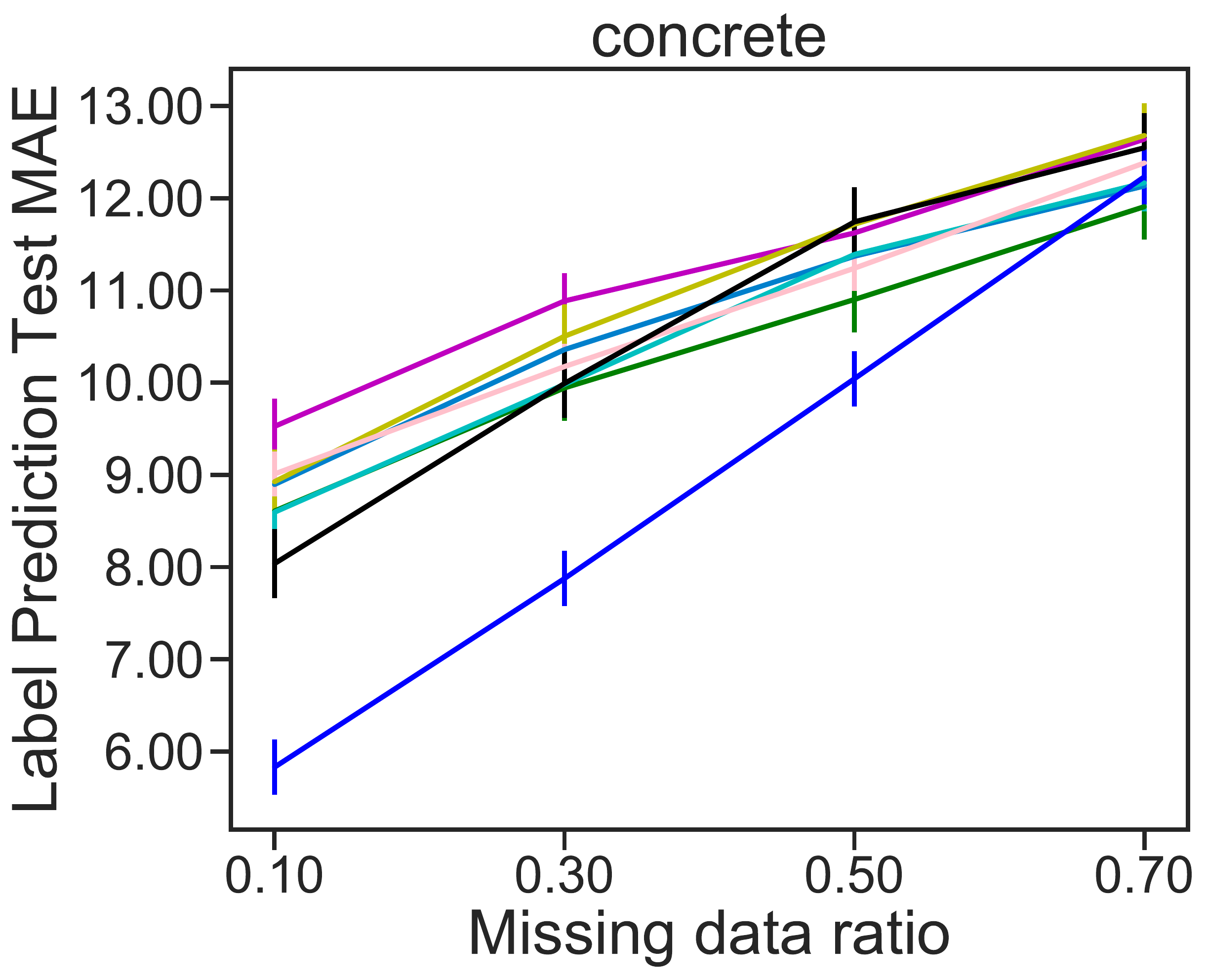}
\includegraphics[width=0.27\columnwidth]{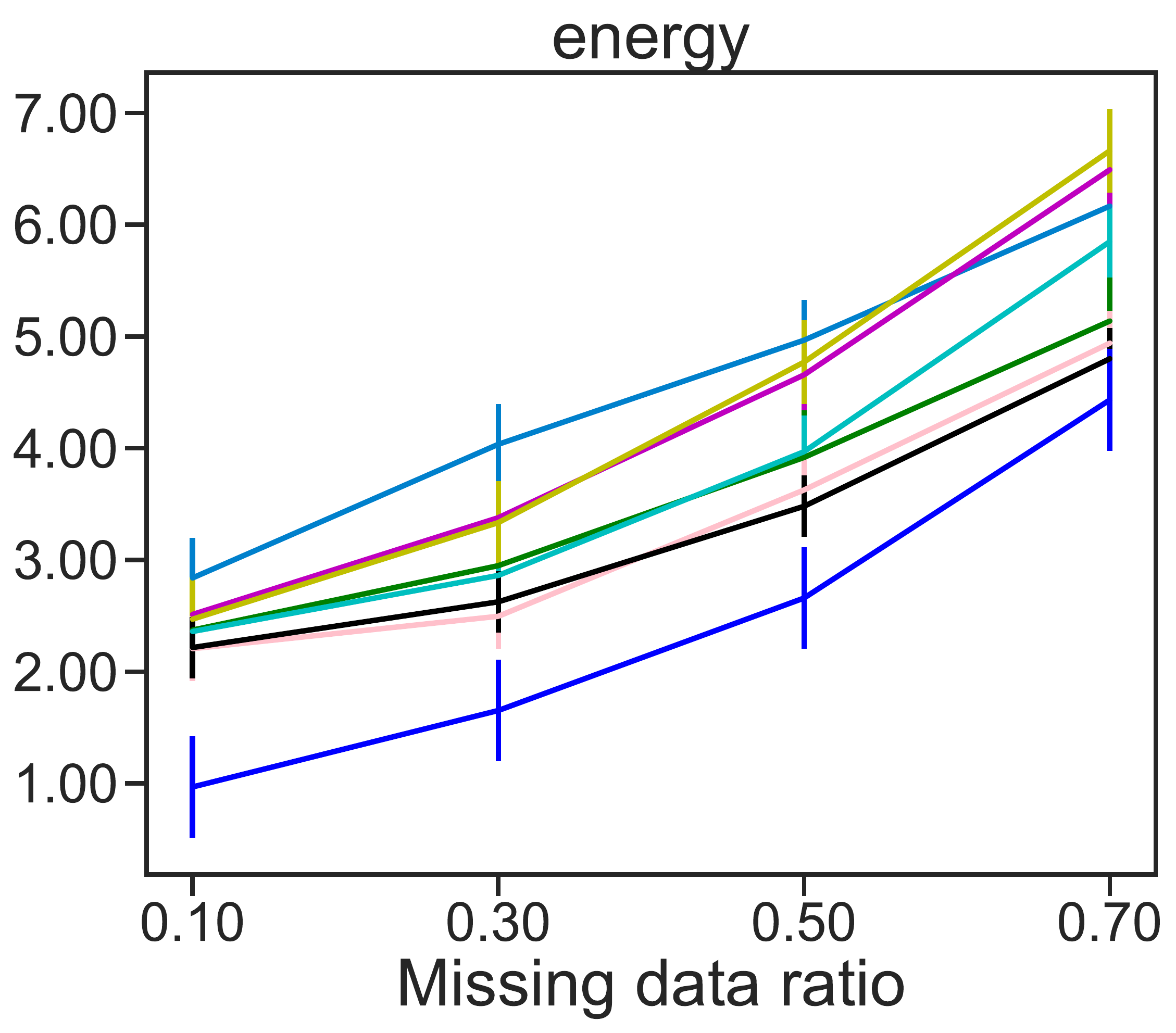}
\includegraphics[width=0.36\columnwidth]{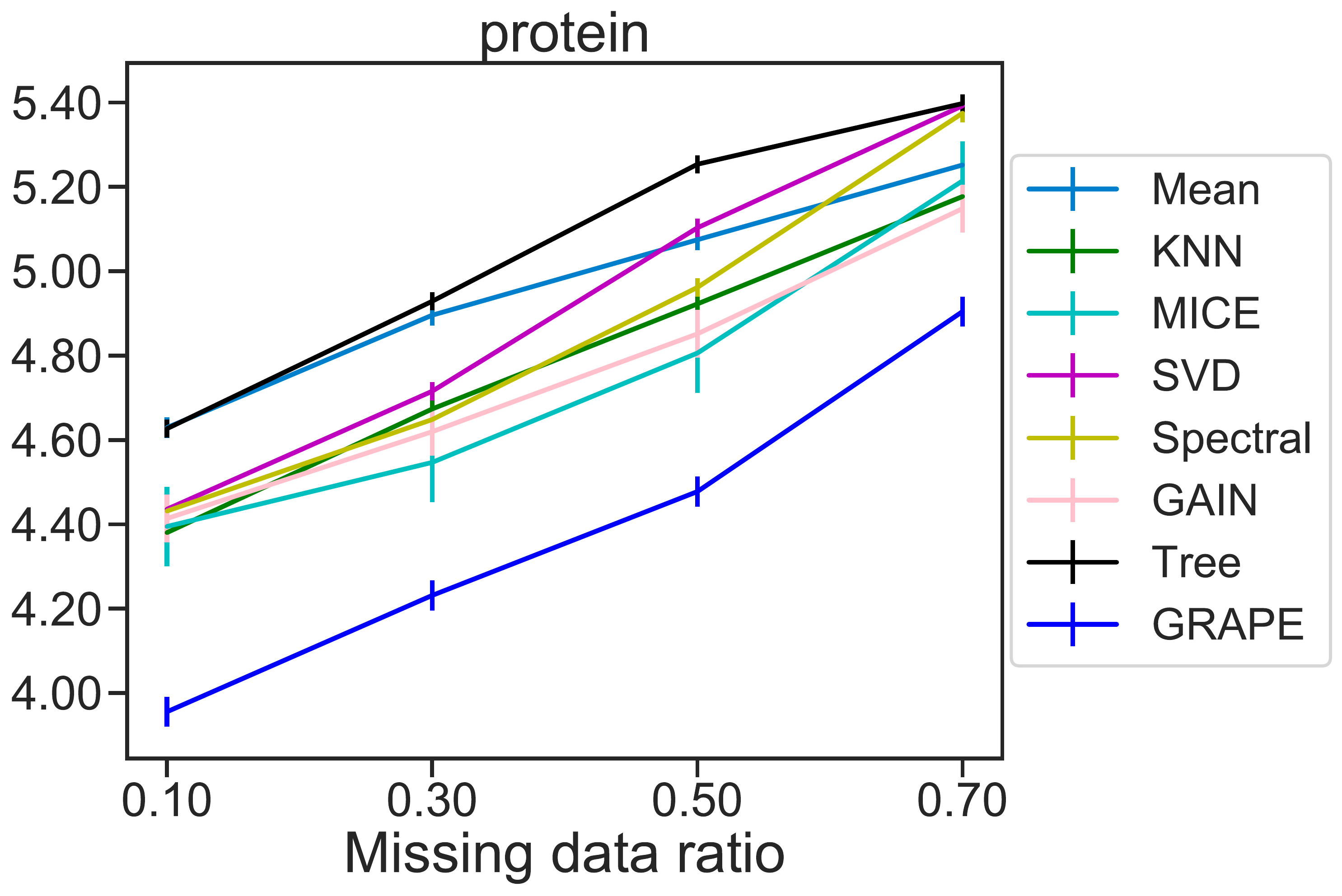}
\caption{Averaged MAE of \emph{feature imputation} (upper) and \emph{label prediction} (lower) with \emph{different missing ratios} over 5 trials. \name yields 12\% lower MAE on imputation and 2\% lower MAE on prediction tasks across different missing data ratios.
}
\vspace{4mm}
\label{fig:mdi_reg_miss}
\end{figure}

\subsection{Label Prediction}
\label{subsec:uci_reg}
\xhdr{Setup}
For label prediction experiments, with the same input graph $\mathcal{G}$, we have an additional label vector $\mathbf{Y}\in \mathbb{R}^n$. 
We randomly split the labels $\mathbf{Y}$ into 70/30\% training and test sets, $\mathbf{Y}_{train}$ and $\mathbf{Y}_{test}$ respectively. 
The training loss is defined as the MSE between the true $\mathbf{Y}_{train}$ and the predicted $\hat{\mathbf{Y}}_{train}$. 
The test metric is calculated based on the MAE between $\mathbf{Y}_{test}$ and $\hat{\mathbf{Y}}_{test}$. For baselines except decision tree, since no end-to-end approach is available, we first impute the data and then do linear regression on the imputed data matrix for predicting $\hat{\mathbf{Y}}$.

\xhdr{Results}
As is shown in \Cref{fig:mdi_reg}, on all datasets except \textsc{naval} and \textsc{wine}, \name has the best performance. On \textsc{wine} dataset, all methods have comparable performance. The fact that the performance of all methods are close to the Mean method indicates that the relation between the labels and observations in \textsc{wine} is relatively simple. For the dataset \textsc{naval}, the imputation errors of all models are very small (both relative to Mean and on absolute value). In this case, a linear regression on the imputed data is enough for label prediction. Across all datasets, \name yields 10\% lower MAE compared with best baselines. The improvement of \name could be explained by two reasons: first, the better handling of missing data with \name where the known information and the missing values are naturally embedded in the graph; and second, the end-to-end training.

\subsection{Robustness against Different Data Missing Levels}
\label{subsec:uci_miss}
\xhdr{Setup}
To examine the robustness of \name with respect to the missing level of the data matrix. We conduct the same experiments as in \Cref{subsec:uci_mdi,subsec:uci_reg} with different missing levels of $r_{miss} \in \{0.1, 0.3, 0.5, 0.7\}$.

\xhdr{Results}
The curves in \Cref{fig:mdi_reg_miss} demonstrate the performance change of all methods as the missing ratio increases. \name yields -8\%, 20\%, 20\%, and 17\% lower MAE on imputation tasks, and -15\%, 10\%, 10\%, and 4\% lower MAE on prediction tasks across all datasets over missing ratios of 0.1, 0.3, 0.5, and 0.7, respectively.
In missing ratio of 0.1, the only baseline that behaves better than \name is KNN. As in this case, the known information is adequate for the nearest-neighbor method to make good predictions. As the missing ratio increases, the prediction becomes harder and the \name's ability to coherently combine all known information becomes more important.

\subsection{Generalization on New Observations}
\label{subsec:uci_generalization}
\xhdr{Setup}
We further investigate the \emph{generalization} ability of \name. Concretely, we examine whether a trained \name can be successfully applied to new observations that are not in the training dataset. A good generalization ability reduces the effort of re-training when there are new observations being recorded after the model is trained.
We randomly divide the $n$ observations in $\mathbf{D}\in \mathbb{R}^{n\times m}$ into two sets, represented as $\mathbf{D}_{train}\in \mathbb{R}^{n_{train}\times m}$ and $\mathbf{D}_{test}\in \mathbb{R}^{n_{test}\times m}$, where $\mathbf{D}_{train}$ and $\mathbf{D}_{test}$ contain 70\% and 30\% of the observations, respectively. The missing rate $r_{miss}$ is at 0.3. 
We construct two graphs $\mathcal{G}_{train}$ and $\mathcal{G}_{test}$ based on $\mathbf{D}_{train}$ and $\mathbf{D}_{test}$, respectively.
We then train \name with $\mathbf{D}_{train}$ and $\mathcal{G}_{train}$ using the same procedure as described in \Cref{subsec:uci_mdi}.
At test time, we directly feed $\mathcal{G}_{test}$ to the trained \name and evaluate its performance on predicting the missing values in $\mathbf{D}_{test}$.
We repeat the same procedure for GAIN where training is also required. For all other baselines, since they do not need to be trained, we directly apply them to impute on $\mathbf{D}_{test}$.

\begin{figure}[t]
\centering
\includegraphics[width=0.9\columnwidth]{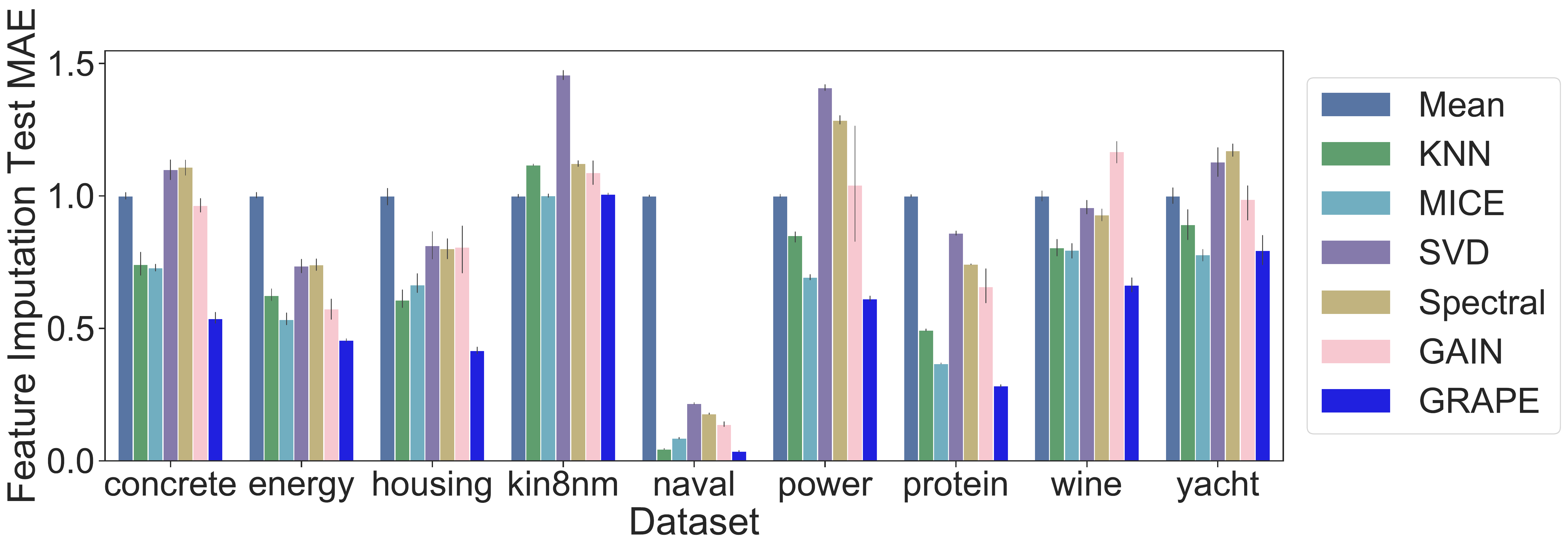}
\caption{Averaged MAE of \emph{feature imputation on unseen data} in UCI datasets over 5 trials. The result is normalized by the average performance of Mean imputation. \name yields 21\% lower MAE compared with best baselines (MICE).
}
\label{fig:mdi_new}
\end{figure}

\xhdr{Results}
As shown in \Cref{fig:mdi_new}, \name yields 21\% lower MAE compared with best baselines (MICE) without being retrained, indicating that our model generalizes seamlessly to unseen observations. Statistical methods have difficulties transferring the knowledge in the training data to new data. While GAIN is able to encode such information in the generator network, it lacks the ability to adapt to observations coming from a different distribution. However, by using a GNN, \name is able to make predictions conditioning on the entire new datasets, and thus capture the distributional changes.

\subsection{Ablation Study}
\label{subsec:edgedropout}

\begin{table}[t]
\setlength\tabcolsep{3pt}
\begin{center}
\caption{\textbf{Ablation study for \name}. Averaged MAE of \name on UCI datasets over 5 trials. Edge dropout (upper) reduces the average MAE by 33\% on feature imputation tasks. $\textsc{Mean}(\cdot)$ is adopted in our implementation. End-to-End training (lower) reduces the average MAE by 19\% on prediction tasks (excluding two outliers).}
\vspace{2mm}
\resizebox{0.9\columnwidth}{!}{
\begin{tabular}{lcccccccccc} 
 \toprule
 & concrete & energy & housing & kin8nm & naval & power & protein & wine & yacht \\ \midrule
Without edge dropout & 0.171& 0.148& 0.104& 0.262& 0.021& 0.192& 0.047& 0.094& 0.204\\ 
\textbf{With edge dropout} & \textbf{0.090} & \textbf{0.136}& \textbf{0.075}& \textbf{0.249}& \textbf{0.008}& \textbf{0.102}& \textbf{0.027}& \textbf{0.063}& \textbf{0.151}\\ \midrule
$\textsc{Sum}(\cdot)$ & 0.094 & 0.143 & 0.078 & 0.277 & 0.024 & 0.134 & 0.040 & 0.069 & 0.154
\\ 
$\textsc{Max}(\cdot)$ & \textbf{0.088} & 0.142 & \textbf{0.074} & 0.252 & \textbf{0.006} & \textbf{0.102} & \textbf{0.024} & \textbf{0.063} & 0.153
\\ 
\textbf{$\textsc{Mean}(\cdot)$}& 0.090 & \textbf{0.136} & 0.075 & \textbf{0.249} & 0.008 & \textbf{0.102} & 0.027 & \textbf{0.063} & \textbf{0.151}
\\ \midrule
Impute then predict & 9.36& 2.59& 3.80& 0.181& \textbf{0.004}& 4.80& 4.48& \textbf{0.524}& 9.02\\ 
\textbf{End-to-End} & \textbf{7.88}& \textbf{1.65}& \textbf{3.39}& \textbf{0.163}& 0.007& \textbf{4.61}& \textbf{4.23}& 0.535& \textbf{4.72}\\ 
 \bottomrule
\end{tabular}
}
\end{center}

\label{tab:abl_study}
\end{table}

\xhdr{Edge dropout} We test the influence of the edge dropout on the performance of \name. We repeat the experiments in \Cref{subsec:uci_mdi} for \name with no edge dropout and the comparison results are shown in \Cref{tab:abl_study}. The edge dropout reduces the test MAE by 33\% on average, which verifies our assumption that using edge dropout could help the model learn to predict unseen edge values.

\xhdr{Aggregation function}
We further investigate how the aggregation function ($\textsc{Sum}(\cdot)$, $\textsc{Max}(\cdot)$, $\textsc{Mean}(\cdot)$) of GNN affects \name's performance.
While $\textsc{Sum}(\cdot)$ is theoretically most expressive, in our setting the degree of a specific node is determined by the number of missing values which is random and unrelated to the missing data task; in contrast, the $\textsc{Mean}(\cdot)$ and $\textsc{Max}(\cdot)$ aggregators are not affected by this inherent randomness of node degree, therefore they perform better.

\xhdr{End-to-end downstream regression}
To show the benefits of using end-to-end training in label prediction, we repeat the experiments in \Cref{subsec:uci_reg} by first using \name to impute the missing data and then perform linear regression on the imputed dataset for node labels (which is the same prediction model as the linear layer used by \name). The results are shown in \Cref{tab:abl_study}.  The end-to-end training gets 19\% less averaged MAE over all datasets except \textsc{naval} and \textsc{wine}. The reason for the two exceptions is similar as described in \Cref{subsec:uci_reg}.

\subsection{Further Discussions}
\xhdr{Scalability}
In our paper, we use UCI datasets as they are widely-used datasets for benchmarking imputation methods, with \emph{both discrete and continuous features}.
\name can easily scale to datasets with thousands of features.
We provide additional results on larger-scale benchmarks, including Flixster (2956 features), Douban (3000 features), and Yahoo (1363 features) in the Appendix.
\name can be modified to scale to even larger datasets.
We can use scalable GNN implementations which have been successfully applied to graphs with billions of edges \cite{ying2018graph,you2019hierarchical}; 
when the number of features is prohibitively large, we can use a trainable embedding matrix to replace one-hot node features.

\xhdr{Applicability of \name}
In the paper, we adopt the most common evaluation regime used in missing data papers, i.e., features are missing completely at random.
\name can be easily applied to other missing data regimes where feature are not missing at random, since \name is fully data-driven.

\xhdr{More intuitions on why \name works}
When a feature matrix does not have missing values, to make downstream label predictions, a reasonable solution will be directly feeding the feature matrix into an MLP. As is discussed in \cite{you2020graph}, an MLP can in fact be viewed as a GNN over a complete graph, where the message function is matrix multiplication.
Under this interpretation, \name extends a simple MLP by allowing it to operate on sparse graphs (\ie, feature matrix with missing values), enabling it for missing feature imputation tasks, and adopting a more complex message computation as we have outlined in Algorithm 1.

\section{Conclusion}
In this work, we propose \name, a framework to coherently understand and solve missing data problems using \emph{graphs}. By formulating the \emph{feature imputation} and \emph{label prediction} tasks as edge-level and node-level predictions on the graph, we are able to train a Graph Neural Network to solve the tasks end-to-end. We further propose to adapt existing GNN structures to handle continuous edge values. Our model shows significant improvement in both tasks compared against state-of-the-art imputation approaches on nine standard UCI datasets. It also generalizes robustly to unseen data points and different data missing ratios. We hope our work will open up new directions on handling missing data problems with graphs.

\section*{Broader Impact}

The problem of missing data arises in almost all practical statistical analyses. The quality of the imputed data influences the reliability of the dataset itself as well as the success of the downstream tasks. Our research provides a new point of view for analysing and handling missing data problems with \emph{graph representations}. There are many benefits to using this framework.
First, different from many existing imputation methods which rely on good heuristics to ensure the performance~\cite{spinelli2019missing}, \name formulates the problem in a natural way without the need of handcrafted features and heuristics. This makes our method ready to use for datasets coming from different domains.
Second, similar to convolutional neural networks~\cite{resnet,vgg}, \name is suitable to serve as a pre-processing module to be connected with downstream task-specific modules. \name could either be pre-trained and fixed or concurrently learned with downstream modules.
Third, \name is general and flexible. There is little limitation on the architecture of the graph neural network as well as the imputation ($\mathbf{O}_{edge}$) and prediction ($\mathbf{O}_{node}$) module. Therefore, researchers can easily plug in domain-specific neural architectures, e.g., BERT \cite{devlin2018bert}, to the design of \name.
Overall, we see exciting opportunities for \name to help researchers handle missing data and thus boost their research.

\section*{Acknowledgments}
We gratefully acknowledge the support of
DARPA under Nos. FA865018C7880 (ASED), N660011924033 (MCS);
ARO under Nos. W911NF-16-1-0342 (MURI), W911NF-16-1-0171 (DURIP);
NSF under Nos. OAC-1835598 (CINES), OAC-1934578 (HDR), CCF-1918940 (Expeditions), IIS-2030477 (RAPID);
Stanford Data Science Initiative, 
Wu Tsai Neurosciences Institute,
Chan Zuckerberg Biohub,
Amazon, Boeing, JPMorgan Chase, Docomo, Hitachi, JD.com, KDDI, NVIDIA, Dell. 
J. L. is a Chan Zuckerberg Biohub investigator.

\bibliography{bibli}
\bibliographystyle{abbrv}

\newpage
\appendix
\section{Additional Details on Baseline Implementation}
For imputation baselines including Mean, KNN, MICE, SVD, and Spectral, we use the implementation provided in the \emph{fancyimpute} package\footnote{\url{https://github.com/iskandr/fancyimpute}}.
For KNN, we use 50 nearest neighbors. For SVD, we set the \textit{rank} equal to $m-1$, where $m$ is the number of features. For MICE, we set the \textit{maximum iteration number} to 3. For Spectral, we found the default heuristic for \textit{shrinkage value} works the best. For a detailed explanation of the meaning of the parameters, we refer readers to the documentation of \emph{fancyimpute} package. The hyper-parameter values are chosen by comparing the average imputation performance over all datasets.
For GAIN, we use the source code released by the authors. All the hyper-parameters are the same as in the source code\footnote{\url{https://github.com/jsyoon0823/GAIN}}.
We use the \emph{rpart} R package for the implementation of the decision tree method.

\section{Running Time Comparison}
Here we report the running clock time for feature imputation of different methods at test time. For Mean, KNN, MICE, SAC, and Spectral, this means the running time of one function call for imputing the entire dataset. For GAIN and \name, this means one forward pass of the network. \Cref{tab:time} shows the averaged running time over 5 different trials with the same setting as described in Section 4.2.

\begin{table}[!htbp]
\setlength\tabcolsep{3pt}
\begin{center}
\caption{Running clock time (second) for feature imputation of different methods at test time.}
\vspace{2mm}
\resizebox{0.95\columnwidth}{!}{
\begin{tabular}{lcccccccccc} 
 \toprule
 & concrete & energy & housing & kin8nm & naval & power & protein & wine & yacht \\ \midrule
Mean & $0.000806$& $0.000922$& $0.000942$& $0.00242$& $0.00596$& $0.00147$& $0.0127$& $0.00121$& $0.00064$\\ 
KNN & $0.225$& $0.134$& $0.0913$& $9.95$& $30.1$& $11.4$& $656$& $0.504$& $0.0268$\\ 
MICE & $0.0294$& $0.0311$& $0.0499$& $0.0749$& $0.256$& $0.0249$& $0.271$& $0.0531$& $0.027$\\ 
SVD & $0.0659$& $0.0192$& $0.0359$& $0.162$& $0.0612$& $0.142$& $0.593$& $0.0564$& $0.0412$\\ 
Spectral & $0.0718$& $0.0565$& $0.0541$& $0.268$& $0.405$& $0.199$& $1.63$& $0.0978$& $0.0311$\\ 
GAIN & $0.0119$& $0.0125$& $0.0131$& $0.017$& $0.0298$& $0.0146$& $0.0457$& $0.0131$& $0.0116$\\ 
\name & $0.0263$& $0.011$& $0.0115$& $0.0874$& $0.259$& $0.0488$& $0.568$& $0.0199$& $0.00438$\\
 \bottomrule
\end{tabular}
}
\end{center}
\label{tab:time}
\end{table}

\section{Comparisons with Additional Baselines}
We additionally provide the comparison results of our method with two other state-of-the-art baselines: missMDA~\cite{josse2016missmda}, a statistical multiple imputation approach, and MIWAE~\cite{mattei2019miwae}, a deep generative model.
We adapt the same setting as in Section 4.1 and the results are shown in \Cref{tab:missMDA_MIWAE}. \name yields the smallest imputation error on all datasets compared with the two other baselines.

\begin{table}[!htbp]
\begin{center}
\caption{Averaged MAE of \emph{feature imputation} on UCI datasets at data missing level of 0.3.}
\vspace{2mm}
\resizebox{0.95\columnwidth}{!}{
\begin{tabular}{cccccccccc}
 \toprule
 &  concrete & energy & housing & kin8nm & naval & power & protein & wine & yacht \\ \midrule
missMDA        & 0.190 & 0.225 & 0.142 & 0.285 & 0.038 & 0.215 & 0.068 & 0.090 & 0.226 
\\ 
MIWAE          & 0.156 & 0.153 & 0.098 & 0.262 & 0.020 & 0.117 & 0.042 & 0.087 & 0.224 
\\
\name & \textbf{0.090} & \textbf{0.136} & \textbf{0.075} & \textbf{0.249} & \textbf{0.008} & \textbf{0.102} & \textbf{0.027} & \textbf{0.063} & \textbf{0.151} \\  \bottomrule
\end{tabular}
}
\end{center}
\label{tab:missMDA_MIWAE}
\end{table}

\section{Experiments on Larger Datasets}
To test the scalability of \name, we perform additional \emph{feature imputation} tests on the Flixter, Douban, and YahooMusic detests with preprocessed subsets and splits provided by \cite{monti2017geometric}. The Flixster dataset has 2341 observations and 2956 features. The Douban dataset has 3000 observations and 3000 features. The YahooMusic dataset has 1357 observations and 1363 features. These datasets only have discrete values. We compare \name with two GNN-based approaches, GC-MC~\cite{berg2017graph} and IGMC~\cite{zhang2019inductive}. The results are shown in \Cref{tab:douban}, where the results of GC-MC and IGMC are provided by \cite{zhang2019inductive}. On all datasets, \name shows a reasonable performance which is better than GC-MC and close to IGMC. Notice that the two baselines are specially designed for discrete matrix completion, where \name is applicable to both continuous and discrete feature values and is general for both feature imputation and label prediction tasks.

\begin{table}[!htpb]
\begin{center}
\caption{RMSE test results on Flixster, Douban, and YahooMusic.}
\label{tab:douban}
\vspace{2mm}
\begin{tabular}{cccc}
\toprule
 &  Flixster & Douban & Yahoo  \\ \midrule
GC-MC  & 0.917 & 0.734 & 20.5 \\ 
IGMC & \textbf{0.872} & \textbf{0.721} & \textbf{19.1} \\ 
Ours & 0.899 & 0.733 & 19.4 \\ \bottomrule
\end{tabular}
\end{center}
\end{table}

\end{document}